\documentclass[sigconf]{acmart}

\usepackage{subcaption}
\AtBeginDocument{%
  \providecommand\BibTeX{{%
    \normalfont B\kern-0.5em{\scshape i\kern-0.25em b}\kern-0.8em\TeX}}}

\setcopyright{acmcopyright}
\copyrightyear{2020} 
\acmYear{2020} 
\setcopyright{acmcopyright}\acmConference[DEEM'20]{International Workshop on Data Management for End-to-End Machine Learning }{June 14, 2020}{Portland, OR, USA}
\acmBooktitle{International Workshop on Data Management for End-to-End Machine Learning (DEEM'20), June 14, 2020, Portland, OR, USA}
\acmPrice{15.00}
\acmDOI{10.1145/3399579.3399869}
\acmISBN{978-1-4503-8023-2/20/06}

\begin{document}

\title{Resilient Neural Forecasting Systems}

\author{Michael Bohlke-Schneider}
\author{Shubham Kapoor}
\author{Tim Januschowski}
\email{bohlkem, kapooshu, tjnsch@amazon.com}
\affiliation{
	Amazon Research
}

\renewcommand{\shortauthors}{Bohlke-Schneider et al.}

\begin{abstract}

Industrial machine learning systems face data challenges that are often under-explored in the academic literature. Common data challenges are data distribution shifts, missing values and anomalies. In this paper, we discuss data challenges and solutions in the context of a Neural Forecasting application on labor planning. We discuss how to make this forecasting system~\emph{resilient} to these data challenges. We address changes in data distribution with a periodic retraining scheme and discuss the critical importance of model stability in this setting. Furthermore, we show how our deep learning model deals with missing values natively without requiring imputation. Finally, we describe how we detect anomalies in the input data and mitigate their effect before they impact the forecasts. This results in a fully autonomous forecasting system that compares favourably to a hybrid system consisting of the algorithm and human overrides.
\end{abstract}


\maketitle

\section{Introduction}

Deploying an industrial machine learning (ML) system to production is a key milestone in development roadmaps of many applied machine learning teams. More often than not, this is the result of long development processes that demands coordination among scientists, software engineers, and business teams. This involves algorithm development, feature engineering and data wrangling, establishing data pipelines, setting up production systems, optimizing the model for latency constraints, and aligning the model performance with business metrics and needs. All of these tasks are non-trivial exercises, especially in combination. 

However, many challenges emerge only~\emph{after} a machine learning system has gone live. In the first phase (the path to production), scientists usually focus on reaching a predictive performance that moves the needle for the respective business case. In the second phase, however, the challenge relies on making the ML system maintain its performance consistently. As real-world data flows into the system, consistent performance can be (at least in part) achieved by addressing data issues to make the ML system \emph{resilient}.

In this paper, we discuss how ML pipelines can be made resilient to common data issues: 1) changes in data distribution (concept drift), 2) missing values in input data, and 3) data anomalies in input that negatively impact the ML algorithm. Although we will discuss these issues in the context of a specific application and used ML algorithms (labor planning using neural forecasting models\footnote{This is short for neural network-based forecasting models.}), the discussed issues and proposed approaches are general enough to be applied to other domains.

This paper is structured as follows. In Section~\ref{sec:labor_planning}, we establish the background of our specific application that we will highlight here: forecasting key inputs for labor planning in the Amazon Customer Fulfillment Network. We present the business case (attendance forecasting) and the algorithm that we use for this task.  Section~\ref{sec:challenges} introduces the specific challenges that we encountered in running ML systems on real-world data from our Fulfillment Centers (FCs) and briefly discusses related work. Section~\ref{sec:solutions} presents our approach in dealing with these challenges. We conclude this paper in Section~\ref{sec:conclusion}. 

\section{Background and Related Work}
\label{sec:labor_planning}

Labor planning, also known as workforce planning, estimates the demand of labor to meet the organization's needs (see e.g.,~\cite{callcenters16,Chapados11,ernst2004annotated}). At Amazon, labor planning is essential to appropriately hire and staff our distribution network. Labor planning must consider several factors, such as projected demand, expected attrition of the workforce, and attendance. Thus, accurate estimates of these factors \emph{in the future} enable accurate labor planning. In this paper, we focus on attendance forecasting, which is one of the key aspects in labor planning. Attendance is defined as the ratio of worked hours and scheduled hours. In principle, the number of worked hours must be smaller than the scheduled hours, as the gap between worked and scheduled hours represents absenteeism i.e. associates not being able to work due to sickness. Thus, the attendance rate is in the interval of $[0, 1]$. 

 At Amazon, we forecast the attendance rate with a neural forecasting model~\cite{flunkert2019deepar}. Neural networks have a long history in industrial applications, both at Amazon and beyond~\cite{flunkert2019deepar,gasthaus2019probabilistic,laptev2017,rangapuram2018deep} and their industrial success has recently been reproduced in academia~\cite{smyl2018m4,makridakis2018m4}. We refer the reader to~\cite{Faloutsos2019,faloutsos2018forecasting} for introductions to forecasting and to~\cite{benidis2020neural} for an overview of the literature.

\subsection{DeepAR \& Labor Planning System}
\label{sec:deepar}

We use DeepAR~\cite{flunkert2019deepar} to forecast attendance in our system. DeepAR is a probabilistic forecasting algorithm based on auto-regressive deep recurrent neural networks. In more detail, DeepAR uses a recurrent neural network to estimate the parameters of parametric probability distributions (e.g., the normal distribution or the beta-distribution). Unlike traditional models like ARIMA~\cite{box1968some} which fits on a individual time series, DeepAR is trained on \emph{all} time series in the dataset and allows for flexible distribution representations~\cite{gasthaus2019probabilistic}.

Learning a global model enables forecasting models to leverage patterns across time series, which are common in industrial applications~\cite{janusch18}. Additionally, by training a global model on similar items, DeepAR can generate forecasts for time series with little or no history.


The system that produces attendance rate forecast is a standard batch-processing ML system set-up. Its main backend components are an ETL data pipeline based on Spark~\cite{Spark2016}. The machine learning model part is based on Amazon SageMaker, and we use the DeepAR implementation available in the same~\cite{Liberty2020}. Recent services like Amazon Forecast abstract away such complexity and alleviate much development work.

\subsection{Data Challenges}
\label{sec:challenges}

In this section, we discuss several data challenges that our system is exposed to and that we consider as common concerns among machine learning practitioners. ~\citet{Polyzotis17tutorial} provides a comprehensive treatment in a 
more general context.

\paragraph{Changes in data distribution:} Data distributions change over time (also known as concept drift) due to seasonal factors or changes in the business or organization. In industrial settings, these data distribution changes are often beyond control of the algorithm owners, which makes the implementation of mitigation strategies inevitable. As the data distribution changes, the predictive performance of the current model deteriorates because it assumed a different data distribution at training. Several approaches to detect and handle concept drift are discussed in the literature (e.g.,~\cite{gama_survey_2014,NIPS2019_8420}). Reacting to data distribution changes requires some adaption mechanism of the model (usually either retraining the model or incrementally updating the model with respect to the latest training examples). 

Model updates are either triggered by an \emph{active} or \emph{passive} approach. In the active case, another algorithm detects whether the data distribution changed and triggers model adaption~\cite{hulten_mining_2001}. In the passive case, the model is updated in a pre-defined schedule or on the latest examples~\cite{widmer_learning_1996}.

\paragraph{Missing values:} Missing values are one of the most common data issues in machine learning~\cite{biessmann_datawig_2019}. Missing values in time series data can occur for several reasons, such as sensor failure for Internet-of-Things (IoT) data or "regular" missing values, for example missing sales data when a store is closed. Missing values can prevent the use of specific algorithms that cannot cope with them. Practitioners often impute missing data with simple statistics (such as means, medians, or modes) or employ statistical and/or machine learning algorithms to fill the missing values~\cite{noauthor_statistical_nodate, batista_analysis_2003, mattei_miwae_2019}.

\paragraph{Anomalies:} Anomalies are data points that fall outside an expected data distribution. In applications like fraud detection, anomalies are the relevant data that needs to be detected. In other cases, anomalies represent unusual (and often unwanted) data that negatively impacts the predictive performance of machine learning systems. Anomaly detection for time series data is an active research field (e.g.,~\cite{ahmad_unsupervised_2017, guha_robust_nodate}). Anomalies can negatively impact a downstream forecasting model. Therefore, ML forecasting systems must detect anomalies and mitigate their effect. 

\section{Resilient Neural Forecasting Systems}
\label{sec:solutions}

\begin{figure*}
	\centering
	\begin{subfigure}[b]{0.4\textwidth}   
		\centering 
		\includegraphics[width=\textwidth]{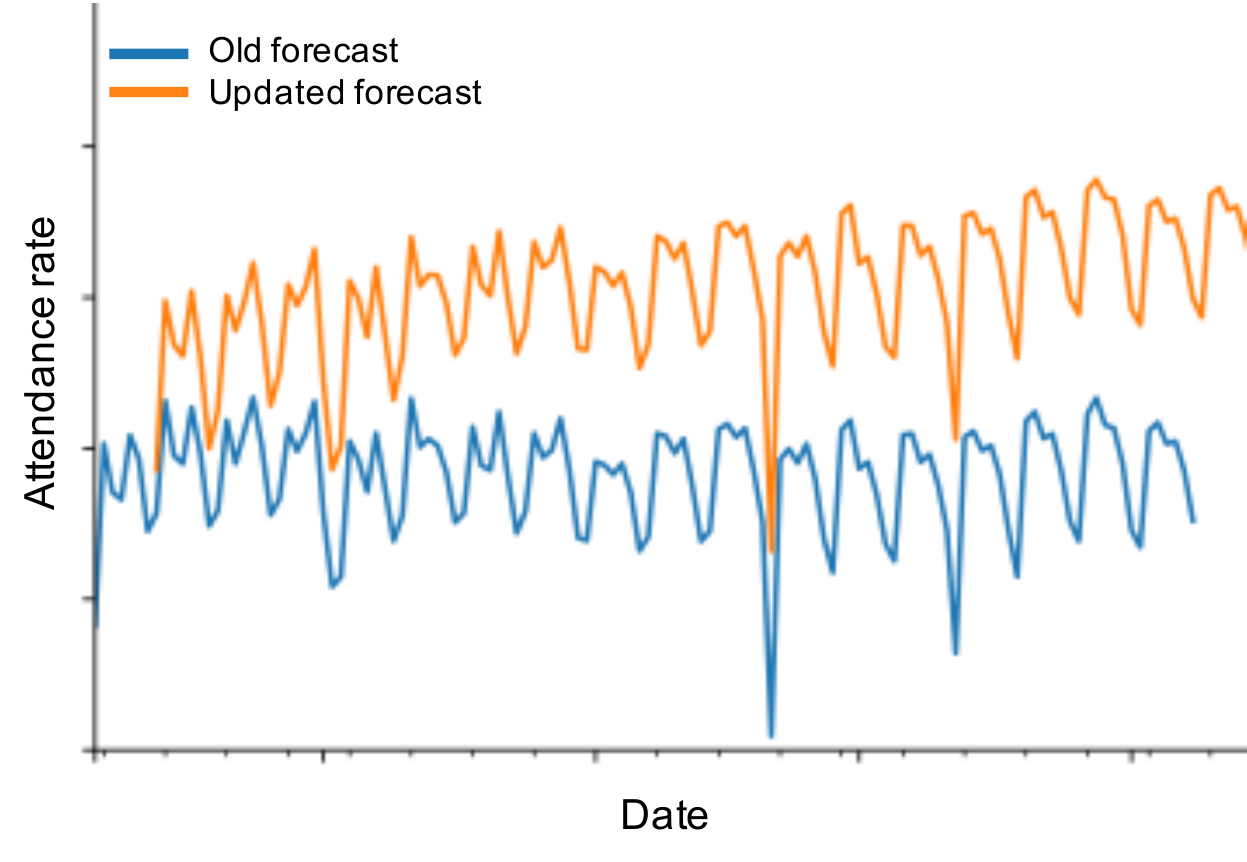}
		\caption[]%
		{{\small No model averaging}}    
		\label{fig:no_ma}
	\end{subfigure}
	\quad
	\begin{subfigure}[b]{0.4\textwidth}   
		\centering 
		\includegraphics[width=\textwidth]{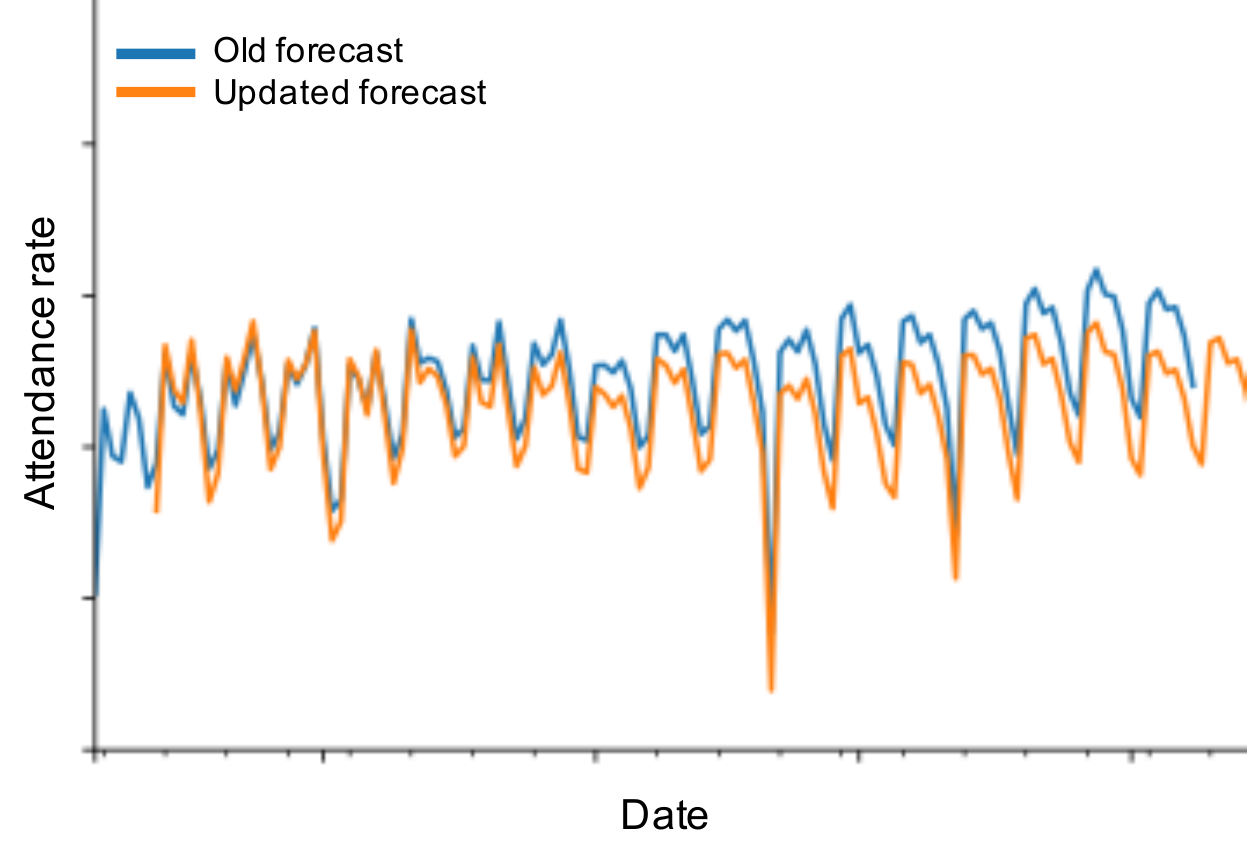}
		\caption[]%
		{{\small With model averaging}}    
		\label{fig:ma}
	\end{subfigure}
	\caption[ The average and standard deviation of critical parameters ]
	{\small Effect of model averaging on our weekly retraining scheme for a selected fulfillment center and two consecutive forecast start dates. We find that without model averaging, DeepAR can be unstable and this can lead to different forecasts upon the weekly updates that are~\emph{not} caused by recent data trends. On average, model averaging improves the spread of forecasted values (average max - min forecasted values on 10 re-runs) by 42.2\%, which we use to measure model stability.} 
	\label{fig:model_averaging}
\end{figure*}

In this section, we illustrate our approaches to the challenges presented in Section~\ref{sec:challenges}. In our system, the implementation of these approaches leads to~\emph{Resilient Neural Forecasting Systems}. We find that our system performs on par or better than human assisted forecasts. Human assisted forecasts allow human operators to override forecasted values with their own estimates. Our automated system improves over human assisted forecasts by 0.18\%. This improvement is small, but we stress that our system is fully automatic and does not require~\emph{any} manual planning, which allows planning teams to focus on other issues.

In this paper, we illustrate a sample set of data and results from experiments with our system. We measure the performance of our system with mean absolute error (MAE) and weighted mean absolute error (wMAE).\footnote{While for more general forecasting problem this is not generally recommended, for data in the [0,1] range, as we have it here, it is an appropriate accuracy measure.}  wMAE is weighted by the size of the shift that the time series represents - for instance, larger shifts have larger weight. This reflects that larger shifts have higher business impact and therefore this is our primary metric.

\subsection{Weekly Retraining Reduces Error but Requires Stable Models} Attendance data is highly seasonal which can be captured by neural forecasting models. Although we routinely produce forecasts with a horizon of eighteen weeks, we also observe distribution changes throughout the years and short-term trends tied to our operational business. To react to these data distribution changes, we rely on a regular interval retraining scheme (weekly retraining). Note that this is somewhat atypical for neural forecasting models which generally require less retraining than classical models (and this counter-balances the often much higher training times of neural forecasting models compared with classical models). In our application, however, we see that including the latest data improves the forecast, on average (wMAE improvement of 3.9\%). This in itself points to data distribution changes being the norm rather than the exception.\footnote{We do not have to retrain neural forecasting models every time, but can, with warm-starting schemes, reduce training times.}

Frequent retraining is a simple, yet effective way to deal with changing data distributions in general, but also has challenges: On the one hand, we would like to capture the natural variation that occurs in the data, which is the primary reason for retraining. On the other hand, machine learning models often have intrinsic variation in their output when retrained. For example, many deep learning algorithms have random weight initializations and are trained by stochastic gradient descent variants. High variation in the output can make retraining schemes impractical because we cannot clearly judge what caused the variation - did the algorithm respond to recent trends and therefore adjusted the forecast? Or are the variations in the forecast caused by low model stability? In the case of forecasting, the forecast itself is just a means to meet a business decision making end. Some business processes cannot be frequently steered (for example, hiring decisions). Thus, low stability models are not aligned with business needs and might reduce the buy in of business partners. Ultimately, models that are not stable will be blocked for production.

 We consider model stability (low variation of output when retraining) as a central attribute. This has several important considerations: 1) Model stability needs to be taken into account when designing experimental work-flows and multiple re-runs need to be the standard setup to estimate mean errors and their corresponding standard error/confidence interval. 2) Model choices and approaches to improve model stability are central algorithmic considerations. 
 
 We stabilize our DeepAR models by using~\emph{model averaging}, which is a simplified parameter averaging technique inspired by Polyak-Ruppert averaging~\cite{polyak_1990,ruppert1988}. Instead of selecting the model with the best loss, we select the $N$ best models and average their parameters (note that this is different from training $N$ models and averaging the outputs of the ensemble). In our attendance forecasting system, model averaging improves the model stability by 42.2\% (Figure~\ref{fig:model_averaging}). Figure~\ref{fig:no_ma} shows an illustrative example of weekly retraining without model averaging and Figure~\ref{fig:ma} with model averaging. 

 Another practical trick to make ourselves more resilient to data shifts is to adapt the scheme to present training examples to DeepAR. These examples are windows in the time series and we can adjust the training sample generation from a uniform scheme where the more distant past is as important as the more recent past, to stratified sampling either according to time (we oversample the more recent history for example) or according to other criteria (important cohorts for which we want to make sure we forecast well for). This naturally helps when a shift in data distribution has occurred.

  \begin{figure*}
 	\centering
 	\begin{subfigure}[b]{0.4\textwidth}   
 		\centering
 		\includegraphics[width=\textwidth]{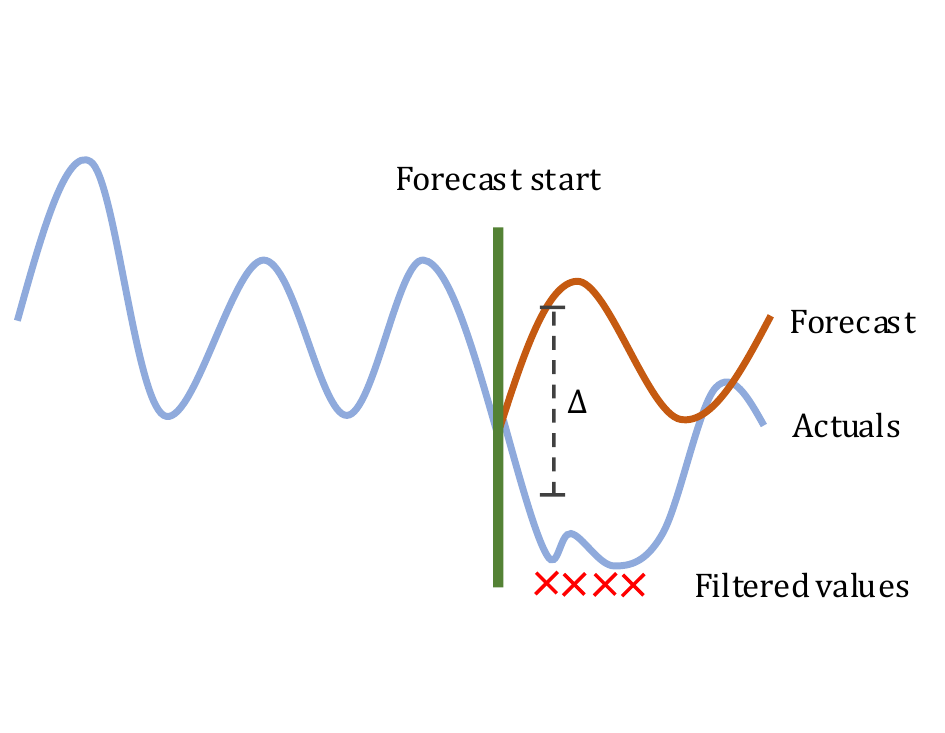}
 		\caption[]%
 		{{\small Schematic illustration of our anomaly detection system}}    
 		\label{fig:anomaly_detection_scheme}
 	\end{subfigure}
 	\quad
 	\begin{subfigure}[b]{0.4\textwidth}   
 		\centering 
 		\includegraphics[width=\textwidth]{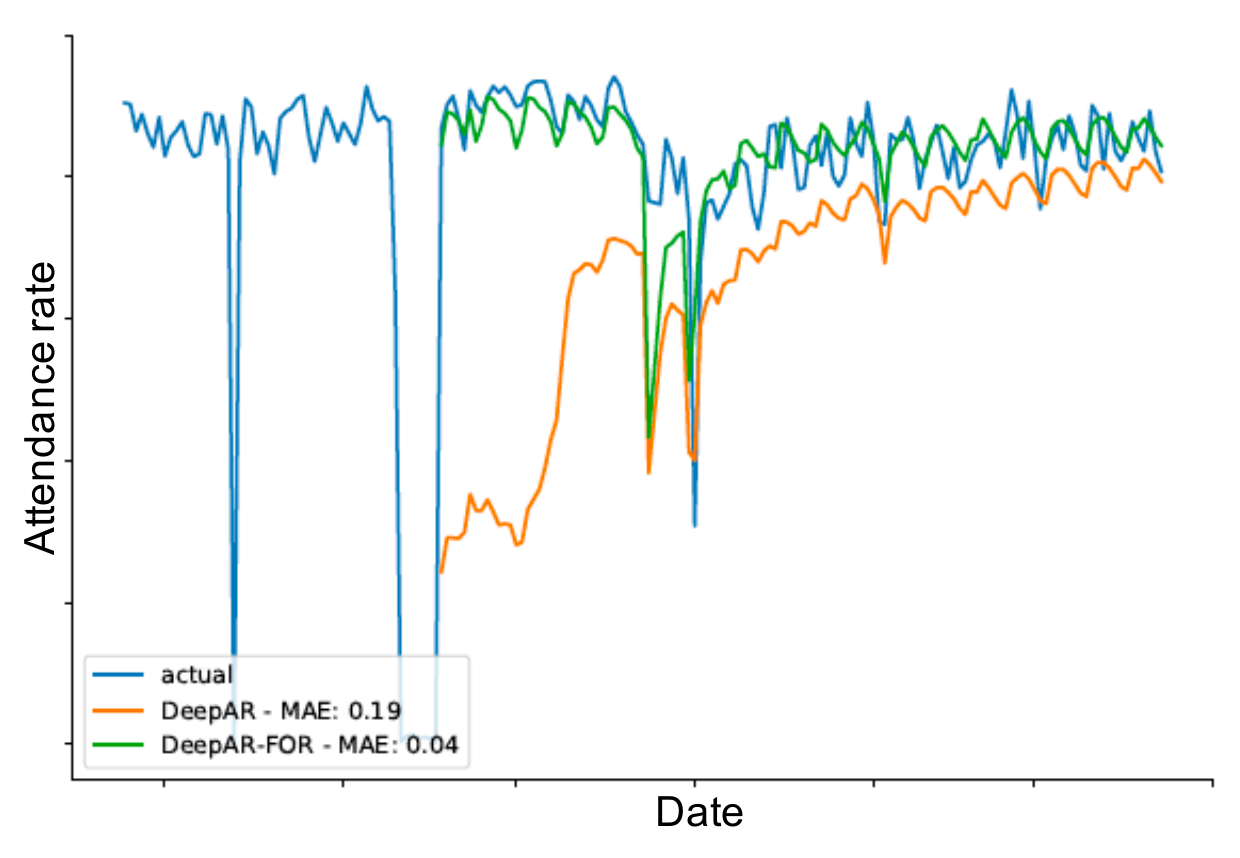}
 		\caption[]%
 		{{\small Effect of our anomaly detection system on an example time series}}    
 		\label{fig:anomaly_detection_example}
 	\end{subfigure}
 	\caption[ The average and standard deviation of critical parameters ]
 	{\small Our anomaly detection system compares the incoming actual values with the forecasted values from the last week. If the difference between forecasted and actual values exceeds a pre-defined $\Delta$, we filter these values and replace them with missing values, which are then handled by our model. In this example, DeepAR without the anomaly detection filter produces poor forecasts (MAE 0.19). Handling these values with the anomaly detection system (DeepAR-FOR), recovers the predictive performance (MAE 0.04).} 
 	\label{fig:no_anomaly_detection}
 \end{figure*}

\subsection{Accepting Missing Values for What They Are} 

In attendance forecasting, missing values occur due to several reasons. For example, some shifts are not worked on particular days (such as weekends), causing a missing value in the time series at weekends. Another source of missing values are seasonal associate shifts that are only worked during specific times of the year (for example, during holiday season). Simple imputation schemes require manual work on the model, at the very least it requires testing multiple imputation methods and monitoring their performance as the data distribution changes.

In the specific case of forecasting, we can leverage that the forecasting model \emph{is} also a model of the time series and therefore naturally lends itself to impute the missing values. DeepAR handles missing values in a principled way: 1) DeepAR masks the loss at missing values to eliminate their effect on the gradient computation, 2) uses missing value indicator features to signal the occurrence of a missing value, and 3) uses the current model parameters to automatically impute the value while training and unrolling the recurrent neural network. This missing value handling is completely automatic and does not require the user to manually impute the values which can lead to unintended side effects.\footnote{For example evaluating forecasts in backtests on imputed values typically leads to wrong conclusions because in such cases forecasting models are favoured that closely resemble the imputation models.} Additionally, dedicated handling of the missing values allows the forecasting models to distinguish missing values that would be in the same range when using manual imputation.

In the next section, we show that this missing value handling enables a principled way of handling anomalies.

\subsection{Anomaly Detection} 

Anomalies are unusually high or low values that can lower the predictive performance of a downstream forecasting system. In our labor planning system, anomalies are introduced by partial or full building closures caused by extreme weather conditions, natural disasters, or power outages. These events arrive as unusually low or even zero attendance in our system, which significantly increase the error of the attendance forecast, if not handled properly~(Figure~\ref{fig:no_anomaly_detection}). 

Again, we leverage that our model is already a model of the time series. In practice, we can compare the forecasted values with the actual data as it becomes available each week. If the difference between forecasted and actual values becomes too large (defined by a manually defined threshold $\Delta$ ), this signals an anomaly and the impact of this data should be mitigated. In our application, we filter the corresponding anomalous value and encode it as a missing value~(Figure~\ref{fig:anomaly_detection_scheme}). Thus, we can let the missing value handling mechanism of DeepAR mitigate the effect of the anomaly. We illustrate this with an incident that closed a fulfillment center for a week. If unchecked, this anomaly arrives as a series of zeros at the forecasting model and leads to a high error forecast (MAE of 0.19). Our anomaly detection system can detect and mitigate the effect of this anomaly via the missing value handling mechanism. Figure~\ref{fig:anomaly_detection_example} shows how our anomaly detection system eliminates the impact of the anomaly and recovers the MAE to 0.04. We also tested simple thresholding heuristics and found that they were less effective in detecting anomalies.

Note that on top of making the forecasting model resilient, this happens automatically and requires no flagging or manual encoding of the anomalous data points. Therefore, this approach is highly scalable.~\footnote{We also find that running the anomaly detection system over the entire data history removes data artefacts and anomalies. We find that removing the anomalies identified this way also reduces the error in our experiments.}  

One shortcoming of this approach is that the forecast cannot distinguish~\emph{anomalies} from~\emph{change points} (qualitative changes in the time series). If change points would be filtered, this would prevent the model to ever learn the changed behavior after the change point! We address this by a simple heuristic: If an anomaly is too long, we assume that this is a change point and we do not filter the values. This requires us to still flag long anomalies manually, but in our case that is a rare event. 

\section{Conclusion}
\label{sec:conclusion}

We presented our work on making a neural forecasting system~\emph{resilient} to common data challenges that arise in industrial machine learning. We find that addressing changes in data distribution, missing values, and anomaly detection results in an automatic system that performs better than human-assisted forecasts. Several key research questions in such systems are still unsolved, such as safely distinguishing anomalies from change points, or how to deal with radical data distribution changes. For example, the COVID-19 pandemic that takes place while writing this paper, has a two-fold effect on forecasting system. First, it violates the core assumption in forecasting that the past is an indication of the future, so it renders forecasts that rely on this assumption as useless. Subject matter experts excel in these situations and outperform automated systems. Second, after the pandemic, another radical data distribution change may occur. Factoring out the impact of the pandemic on many real-world datasets used as training sets for machine learning algorithms will then be another (very welcome) challenge.

\section{Acknowledgements}

We thank our engineering and business partners at Amazon for their cooperation. We thank Alex Kim for comments on this manuscript.

\bibliographystyle{ACM-Reference-Format}
\bibliography{labor_planning_deem_workshop}

\end{document}